\documentclass{article}

\usepackage{microtype}
\usepackage{graphicx}
\usepackage{booktabs} % for professional tables

\usepackage{hyperref}

\usepackage[accepted]{icml2023}

\usepackage{amsmath}
\usepackage{amssymb}
\usepackage{mathtools}
\usepackage{amsthm}
\usepackage{subcaption}
\usepackage{caption}
\usepackage{multirow}
\usepackage{slashbox}
\usepackage{bm}
\usepackage[capitalize,noabbrev]{cleveref}

\theoremstyle{plain}

\theoremstyle{definition}

\theoremstyle{remark}

\usepackage[textsize=tiny]{todonotes}

\icmltitlerunning{Is ReLU Adversarially Robust?}

\begin{document}

\twocolumn[
\icmltitle{Is ReLU Adversarially Robust?}




\begin{icmlauthorlist}
\icmlauthor{Korn Sooksatra}{1}
\icmlauthor{Greg Hamerly}{1}
\icmlauthor{Pablo Rivas}{1}
\end{icmlauthorlist}

\icmlaffiliation{1}{Department of Computer Science, Baylor University, Waco, TX}

\icmlcorrespondingauthor{Korn Sooksatra}{korn\_sooksatra1@baylor.edu}
\icmlcorrespondingauthor{Greg Hamerly}{greg\_hamerly@baylor.edu}
\icmlcorrespondingauthor{Pablo Rivas}{pablo\_rivas@baylor.edu}

\icmlkeywords{Machine Learning, ICML}

\vskip 0.3in
]



\printAffiliationsAndNotice{\icmlEqualContribution} % otherwise use the standard text.

\begin{abstract}
The efficacy of deep learning models has been called into question by the presence of adversarial examples. Addressing the vulnerability of deep learning models to adversarial examples is crucial for ensuring their continued development and deployment. In this work, we focus on the role of rectified linear unit (ReLU) activation functions in the generation of adversarial examples. ReLU functions are commonly used in deep learning models because they facilitate the training process. However, our empirical analysis demonstrates that ReLU functions are not robust against adversarial examples. We propose a modified version of the ReLU function, which improves robustness against adversarial examples. Our results are supported by an experiment, which confirms the effectiveness of our proposed modification. Additionally, we demonstrate that applying adversarial training to our customized model further enhances its robustness compared to a general model.
\end{abstract}

\section{Introduction}

One of deep learning models' significant challenges is their vulnerability to tiny, imperceptible perturbations embedded in their input, known as adversarial examples. An attacker~\cite{goodfellow2014explaining, madry2017towards, sooksatra2022evaluation, sooksatra2021enhancing, kurakin2016adversarial, papernot2016limitations, papernot2016transferability, papernot2017practical, ilyas2018black, gong2017adversarial, kos2018adversarial, kurakin2018adversarial, xiao2018generating, hendrycks2021natural, luo2018towards, zhao2017generating, croce2020reliable} explicitly crafts these perturbations to cause the model to make a wrong prediction.

Several defense mechanisms~\cite{goodfellow2014explaining, madry2017towards, wong2020fast, tramer2017ensemble, zhang2019theoretically, wang2019improving, shafahi2019adversarial, rakin2018defend, lin2019defensive} have been proposed to address this issue, including adversarial training, input preprocessing, and network architecture modification. However, few have focused on fixing the general activation functions (such as Sigmoid, Tanh, and ReLU), which cause adversarial examples. ReLU activation functions are widely used in deep-learning models because they accelerate the training process and address the vanishing gradient problem. However, they also make the models vulnerable to adversarial examples, allowing many tiny perturbations to enlarge themselves over multiple layers in the models.

In this work, inspired by ReLU6 activation functions \cite{sandler2018mobilenetv2}, we propose a solution by experimenting with and customizing ReLU activation functions by capping them. Our results indicate that the models become more robust against adversarial examples when their max values are set. Additionally, we found that the models are even more robust when the max values are decreased. However, this technique is limited to small-scale datasets (such as MNIST~\cite{deng2012mnist}) as it does not address the vanishing gradient problem for medium and large-scale datasets (such as CIFAR10~\cite{krizhevsky2009learning} and Imagenet~\cite{5206848}), which require additional techniques. Our work aims to provide a new perspective on improving the robustness of deep-learning models and making them more trustworthy in practical scenarios.

This paper is organized as follows: Section~\ref{sec:related_works} explores and describes the works that focus on improving adversarial robustness; Section~\ref{sec:relu} shows that the problem of ReLU functions and describes one way to customize them to mitigate the problem; Section~\ref{sec:size} demonstrates how the size of a layer with customized ReLU functions affects the robustness; Section~\ref{sec:order} demonstrates how the order of a layer with customized ReLU functions affects the robustness; Section~\ref{sec:zero_gradient} explains the experiment to support our claim regarding our customized functions and demonstrates its result; Section~\ref{sec:map} discusses how to compute sensitivity map of an image to show that capping ReLU functions leads to robustness; Section~\ref{sec:adversarial_training} shows the effect of adversarial training on a model with our customized functions compared to the one with general functions; Section~\ref{sec:conclusion} concludes our work and describes the limitation of our solution and future works.

\section{Related Works \label{sec:related_works}}

In recent years, a significant amount of research has been dedicated to addressing the vulnerability of deep-learning models to adversarial examples. A wide range of defensive mechanisms has been proposed to promote the robustness of these models, which can broadly be categorized into training and architectural solutions.

The training solution approach aims to improve the robustness of deep-learning models by proposing alternative ways to train a classifier. This can include techniques such as adversarial training, in which the model is trained on adversarial examples to improve its robustness against them.

On the other hand, the architectural solution seeks to alter various parts of the classifier's network to increase its robustness. This can include techniques such as network architecture modification and activation function customization. In this section, we will discuss the existing works within these two categories of solutions, providing an in-depth understanding of the current state of the art in adversarial robustness.

\subsection{Adversarial Training}
\citet{goodfellow2014explaining} proposed an adversarial robustness evaluation called Fast Gradient Sign Method (FGSM). Then, they augmented the adversarial examples from FGSM to the training dataset and retrained a classifier. As a result, the classifier was robust against FGSM. Later, in 2017,~\citet{madry2017towards} suggested using Projected Gradient Descent (PGD)~\cite{madry2017towards} to estimate the inner maximization of adversarial training instead of FGSM because PGD was a stronger attack than FGSM. Also, a classifier retrained with adversarial examples from PGD could be robust against $L_\infty$ attacks (e.g., FGSM and PGD) and $L_2$ attacks (e.g., Carlini and Wagner attack~\cite{carlini2017towards}). \citet{tramer2017ensemble} mentioned that the adversarial training relying on adversarial examples from one classifier does not provide robustness against black-box attacks. Therefore, they used adversarial examples from several classifiers to retrain a classifier to improve its robustness against black-box attacks. In 2019, \citet{zhang2019theoretically} split the objective function for the training process into two terms. The first term was for accuracy, and the second was for robustness. Hence, their training scheme could find the tradeoff between accuracy and robustness. \citet{wang2019improving} improved the adversarial training by emphasizing misclassified training samples. This method led to a more robust classifier than the one trained with adversarial training that did not differentiate the misclassified and correctly classified examples. Furthermore, \citet{shafahi2019adversarial} utilized the gradients from the natural training to compute adversarial perturbations for adversarial training. However, these perturbations could be used only for FGSM. Hence, to make a classifier as robust as using PGD, they needed to train the classifier for much more epochs than general adversarial training. Therefore, in 2020, \citet{wong2020fast} showed that FGSM was good enough to make a classifier robust; thus, we did not need to train the classifier for as many epochs as the work in~\cite{shafahi2019adversarial} did.

\subsection{Architectural solution}
Only a few existing works proposed solutions by altering a classifier's architecture. In 2018, \citet{rakin2018defend} quantized activation functions in a classifier to eliminate adversarial perturbations. They also showed that dynamic quantization could make the classifier more robust than fixing the thresholds of quantization. Later, in 2019, \citet{lin2019defensive} improved this approach by adding a regularization term that could indicate the Lipschitz constant. Nonetheless, minimizing the constant was intractable; hence, the term made the covariance matrix of each layer's weight close to the identity matrix. However, these quantization approaches negatively affected the training process due to the vanishing gradient problem. 

To the best of our knowledge, our work is the first that customizes the activation functions for robustness and reduces the negative impact on the training process. For example, a previous work quantizes activation functions in a model to improve its robustness; however, this technique leaves only zero gradients in the activation functions and harms a gradient-based training process, such as gradient descent. Our work instead leaves some areas of the functions to be differentiable.

\section{Flaws and Modification of ReLU Functions \label{sec:relu}} 
% Unlike Sigmoid and Tanh activation functions, ReLU activation functions are friendly to backpropagation because they have many spaces for gradient computation. Therefore, it is a solution to avoid the vanishing gradient problem. However, they are one of the weaknesses in deep learning models against adversarial examples since tiny perturbations in inputs can be enlarged over the hidden layers. Thus, these tiny perturbations can result in a huge difference in the output layer.

ReLU activation functions have been widely used in deep-learning models due to their ability to accelerate the training process and address the vanishing gradient problem. Unlike Sigmoid and Tanh activation functions, ReLU activation functions have many spaces for gradient computation, making them more friendly to backpropagation. However, this property that makes ReLU functions worthwhile makes them weak in deep-learning models regarding adversarial examples. Because ReLU functions allow many tiny perturbations in inputs to enlarge themselves over the hidden layers, these tiny perturbations can result in a significant difference in the output layer, making the model vulnerable to adversarial examples.

\subsection{Enlarged Perturbations}

\begin{figure*}[h!]
     \centering
     \begin{subfigure}[b]{0.23\textwidth}
         \centering
         \includegraphics[width=\textwidth]{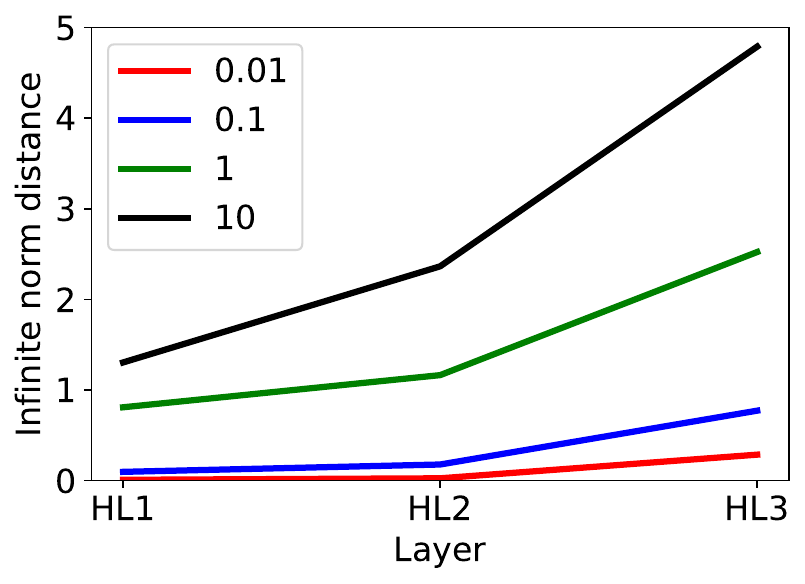}
         \label{fig:growing_perturbation_mnist_linfinite_cap1}
     \end{subfigure}
     \hfill
     \begin{subfigure}[b]{0.23\textwidth}
         \centering
         \includegraphics[width=\textwidth]{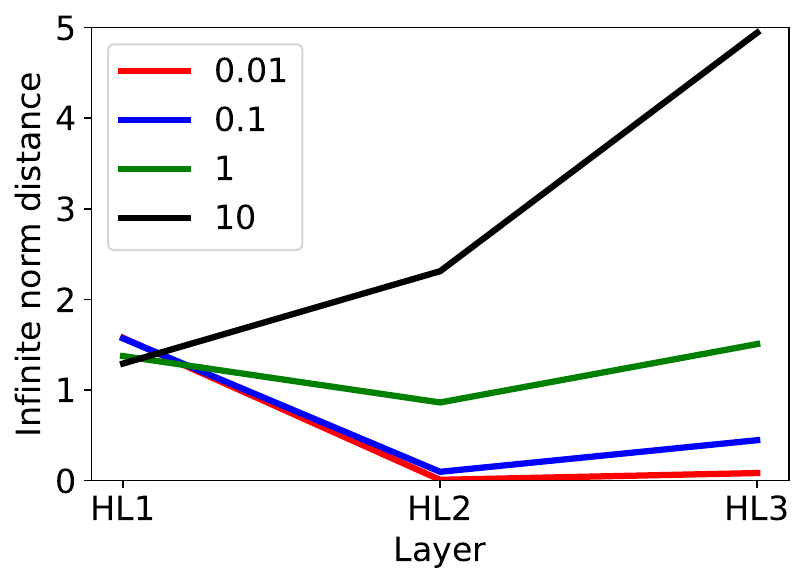}
         \label{fig:growing_perturbation_mnist_linfinite_cap2}
     \end{subfigure}
     \hfill
     \begin{subfigure}[b]{0.23\textwidth}
         \centering
         \includegraphics[width=\textwidth]{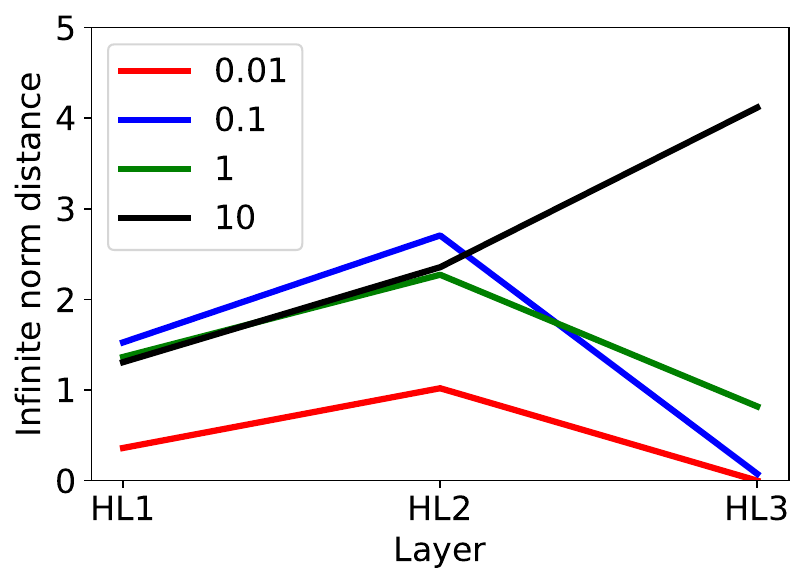}
         \label{fig:growing_perturbation_mnist_linfinite_cap3}
     \end{subfigure}
     \hfill
     \begin{subfigure}[b]{0.23\textwidth}
         \centering
         \includegraphics[width=\textwidth]{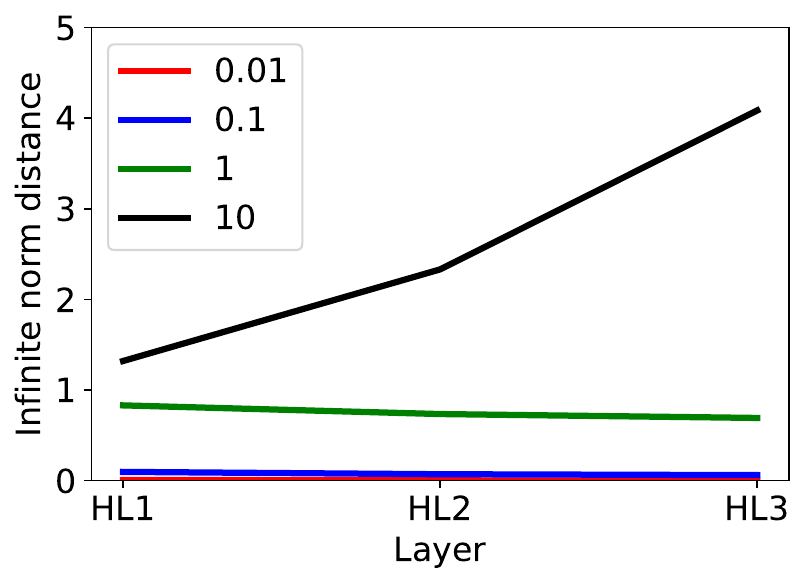}
         \label{fig:growing_perturbation_mnist_linfinite_cap123}
     \end{subfigure}
     \hfill
     \begin{subfigure}[b]{0.23\textwidth}
         \centering
         \includegraphics[width=\textwidth]{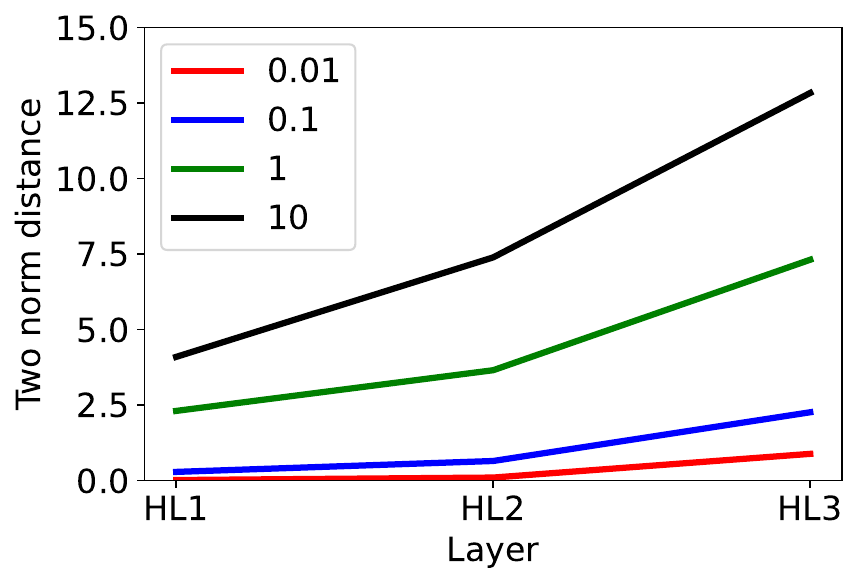}
         \caption{Cap the first hidden layer.}
         \label{fig:growing_perturbation_mnist_l2_cap1}
     \end{subfigure}
     \hfill
     \begin{subfigure}[b]{0.23\textwidth}
         \centering
         \includegraphics[width=\textwidth]{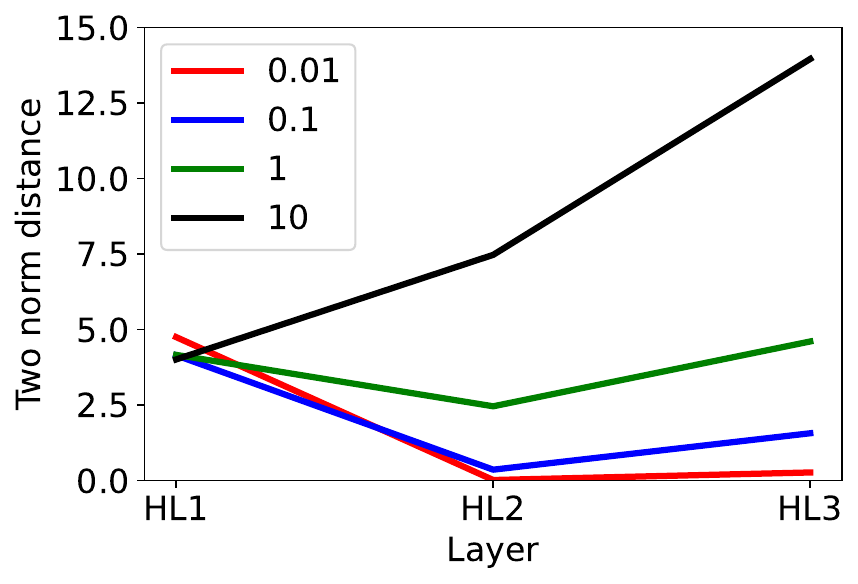}
         \caption{Cap the second hidden layer.}
         \label{fig:growing_perturbation_mnist_l2_cap2}
     \end{subfigure}
     \hfill
     \begin{subfigure}[b]{0.23\textwidth}
         \centering
         \includegraphics[width=\textwidth]{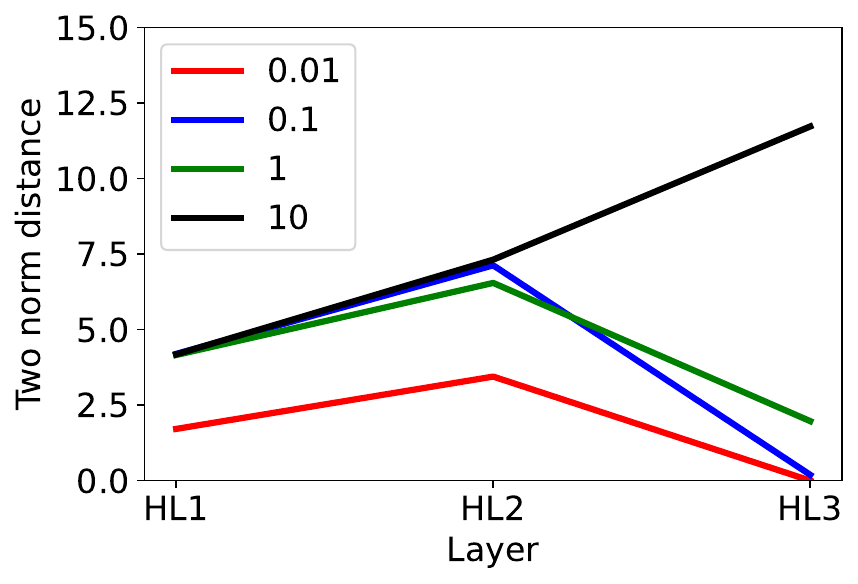}
         \caption{Cap the third hidden layer.}
         \label{fig:growing_perturbation_mnist_l2_cap3}
     \end{subfigure}
     \hfill
     \begin{subfigure}[b]{0.23\textwidth}
         \centering
         \includegraphics[width=\textwidth]{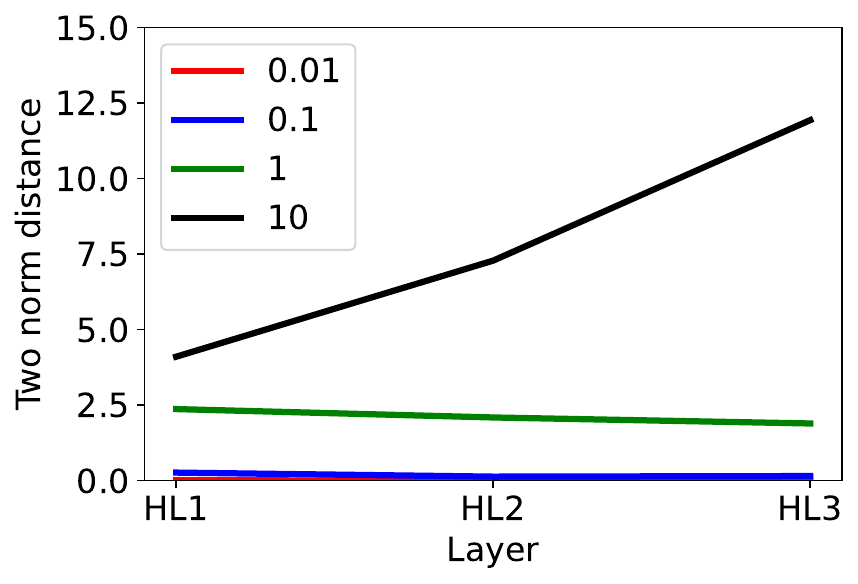}
         \caption{Cap all the hidden layers.}
         \label{fig:growing_perturbation_mnist_l2_cap123}
     \end{subfigure}
        \caption{The distance between each hidden layer's outputs resulted from passing clean samples and adversarial examples. Note that the top row shows the $L_\infty$ distance and the bottom row shows the $L_2$ distance.}
        \label{fig:growing_perturbation_mnist}
\end{figure*}

Fortunately, we found that capping ReLU activation functions can stop the perturbations from growing over the layers. Therefore, we construct an experiment to show the growing perturbations in the hidden layers for various max values. We use the MNIST dataset~\cite{deng2012mnist} and train its classifier consisting of three dense hidden layers whose sizes are $392$, $196$, and $98$. After that, we utilize Projected Gradient Descent (PGD) attack on the classifier and the test dataset with the perturbation bound of $20/256$, step size of $2/256$ and the max iteration of $20$. Then, we obtain adversarial examples. Next, we train other classifiers and cap different hidden layers (i.e., the first hidden layer (HL1), the second hidden layer (HL2), the third hidden layer (HL3) and all the hidden layers (HL123)). Also, we cap them with diverse max values (i.e., $0.01$, $0.1$, $1$, $10$, and $100$). Figure~\ref{fig:growing_perturbation_mnist} demonstrates that ReLU functions with high max values allow perturbations to become huge over the layers. On the other hand, capping them with low values can mitigate such an effect. Further, it is intuitive that the difference significantly goes down at the layer capped, as seen in the figure. 

\begin{figure}[h!]
    \centering
    \includegraphics[width=.35\textwidth]{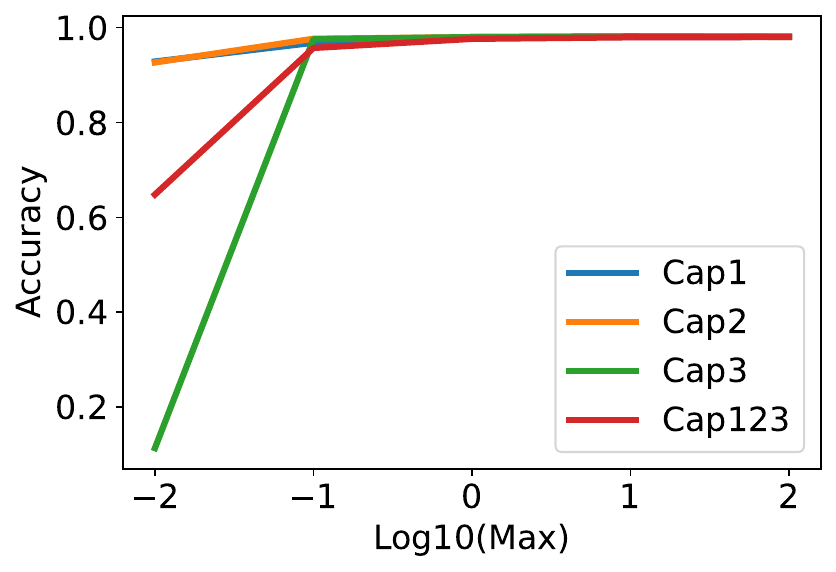}
    \caption{Accuracy achieved by classifiers with different capped hidden layers and max values on MNIST test dataset.}
    \label{fig:growing_perturbation_mnist_acc}
\end{figure}

Although capping ReLU functions reduces the growth of perturbed values that may significantly alter the output, we found that when we set the max value to be very low, the classifier would underfit the dataset due to the vanishing gradient problem, as demonstrated in Figure~\ref{fig:growing_perturbation_mnist_acc}. Also, capping all the hidden layers achieved a slightly lower performance than capping only one hidden layer, as 
 seen in the figure when the max value is $0.1$. Therefore, in this phenomenon, there is a tradeoff between the network's ability to be trained and its sensitivity to tiny perturbations.

% \begin{table*}[h!]
%     \centering
%     \begin{tabular}{|c|ccc|ccc|ccc|ccc|}
%         \hline
%        \multirow{2}{*}{\backslashbox{Capping}{$L_\infty$}}  & \multicolumn{3}{c|}{Max = 0.01} & \multicolumn{3}{c|}{Max = 0.1} & \multicolumn{3}{c|}{Max = 1} & \multicolumn{3}{c|}{Max = 10} \\
%        \cline{2-13}
%         & HL1 & HL2 & HL3 & HL1 & HL2 & HL3 & HL1 & HL2 & HL3 & HL1 & HL2 & HL3\\
%         \hline
%         HL1 & 0.01 & 0.03 & 0.29 & 0.10 & 0.18 & 0.77 & 0.81 & 1.16 & 2.52 & 1.30 & 2.37 & 4.79\\
%         HL2 & 1.58 & 0.01 & 0.08 & 1.57 & 0.10 & 0.45 & 1.38 & 0.86 & 1.51 & 1.29 & 2.31 & 4.94\\
%         HL3 & 0.36 & 1.02 & 0.00 & 1.53 & 2.71 & 0.08 & 1.37 & 2.27 & 0.82 & 1.31 & 2.36 & 4.12\\
%         HL123 & 0.01 & 0.00 & 0.00 & 0.10 & 0.07 & 0.06 & 0.83 & 0.74 & 0.69 & 1.32 & 2.33 & 4.09 \\
%         \hline
%     \end{tabular}
%     \caption{The difference in $L_{\infty}$ between the outputs of each hidden layer where the inputs are clean samples and adversarial examples from another network. Note that HL1 denotes the first hidden layer, HL2 denotes the second hidden layer, HL3 denotes the third hidden layer and HL123 denotes all the hidden layers.}
%     \label{tab:my_label}
% \end{table*}

\subsection{Capped ReLU Function}
We show that the capped ReLU function can effectively control the enlarged perturbations. This section shows the formal definition of the function. 

A capped ReLU function is a general ReLU function capped with a value. Hence, we can formulate this function as
\begin{equation}
    a(z, \beta) = \max(0, \min(z, \beta))
\end{equation}
where $z$ is the function's input and $\beta$ is a max value that caps the function. As seen in Figure~\ref{fig:growing_perturbation_mnist_l2_cap1}, reducing $\beta$ can control the growing perturbations efficiently.

\subsection{Sigmoid and Tanh Activation}

\begin{figure}[h!]
    \centering
    \includegraphics[width=.35\textwidth]{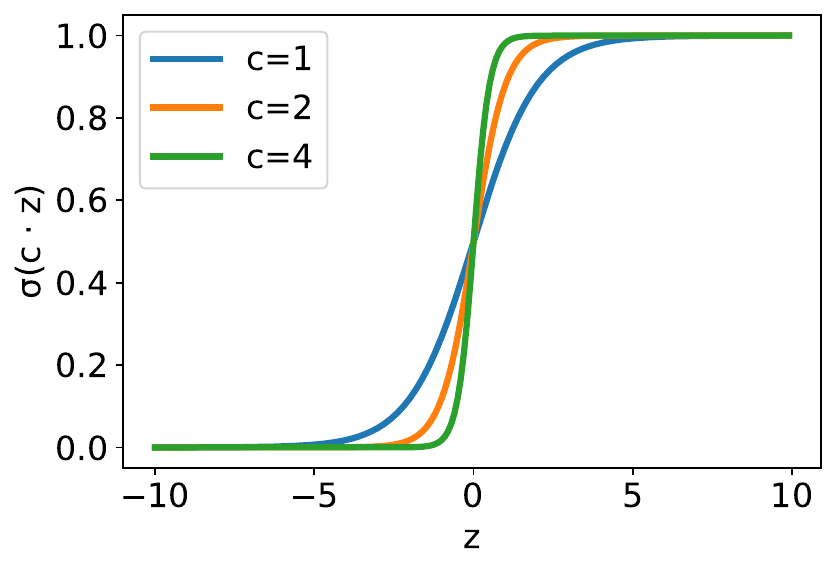}
    \caption{Sigmoid functions with different constant $c$.}
    \label{fig:sigmoid}
\end{figure}

Sigmoid and Tanh activation functions can be good candidates to provide adversarial robustness since their values have the highest and lowest values. However, the spaces from their minimum to maximum values are too wide to stop adversarial perturbations from enlarging themselves over layers. One way to narrow the gap is by multiplying their arguments with some constants. For example, let $\sigma(z)$ be a Sigmoid or Tanh function and $z$ be the result after a layer before this activation. We can narrow it by multiplying its argument with a real value $c$, where $c > 1$. Hence, the function becomes $\sigma(c \cdot z)$. Figure~\ref{fig:sigmoid} shows that increasing $c$ can narrow the function. 

Nevertheless, the parameters of the corresponding layer are a part of the computation of $z$. Thus, the function can become wide when the training reduces those parameters. Then, Sigmoid or Tanh activation functions cannot be used to eliminate adversarial perturbations, unlike ReLU functions.

\section{Effect of Capped Layer's Size on Robustness \label{sec:size}}
In this section, we want to know whether the width of the layer (narrow or wide) containing the capped neurons has an effect on robustness. Therefore, we conducted the following experiment. 
\subsection{Experimental Explanation and Setting}

First, we trained a classifier with some layers capped with an initial max value to control the value after the ReLU function. Then, we evaluated this classifier in terms of accuracy on clean test samples (i.e., standard accuracy) and adversarial examples (i.e., robust accuracy) and the success rate of an attack. However, we cannot rely only on the initial max value because it may not make the classifier the most robust. Therefore, we evaluate this classifier with several max values (i.e., from $0.01$ to $0.15$ in our experiments) to determine the max value that promotes robustness and does not sacrifice much standard accuracy. We used the MNIST dataset for this experiment. Also, we created two kinds of classifiers: a ``general'' two-hidden-layer dense network and a ``reversed'' two-hidden-layer dense network. The former consists of an input layer, a $392$-neuron layer with ReLU activation, a $196$-neuron layer with ReLU activation, and the output layer with Softmax activation. In the latter, only the hidden layers are swapped. Further, the attack we used for this experiment is Projected Gradient Descent (PGD)~\cite{madry2017towards} because it is one of the strong attacks and is widely used for adversarial robustness evaluation.

\subsection{Results}
\begin{figure*}[h!]
     \centering
     \begin{subfigure}[b]{0.3\textwidth}
         \centering
         \includegraphics[width=\textwidth]{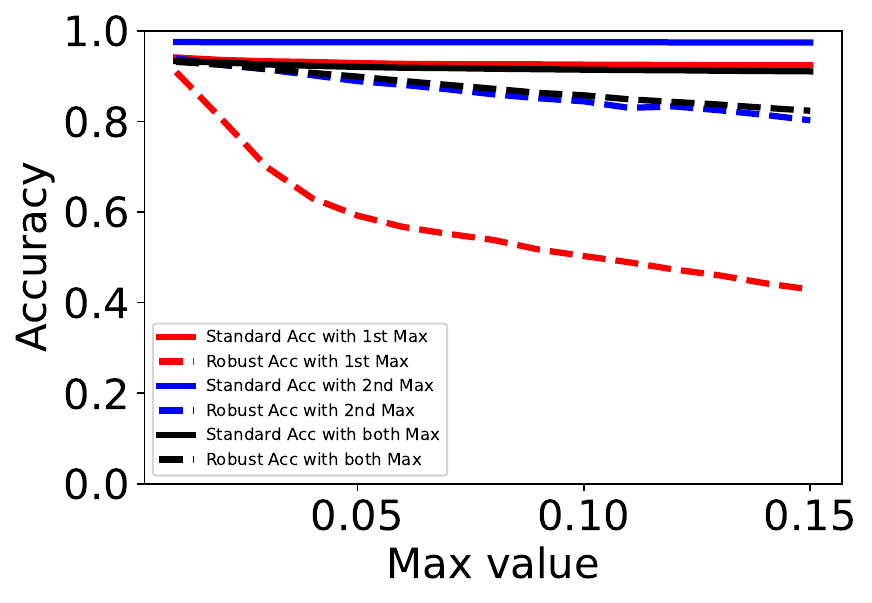}
         \label{fig:0.01_acc_two_hidden}
     \end{subfigure}
     \hfill
     \begin{subfigure}[b]{0.3\textwidth}
         \centering
         \includegraphics[width=\textwidth]{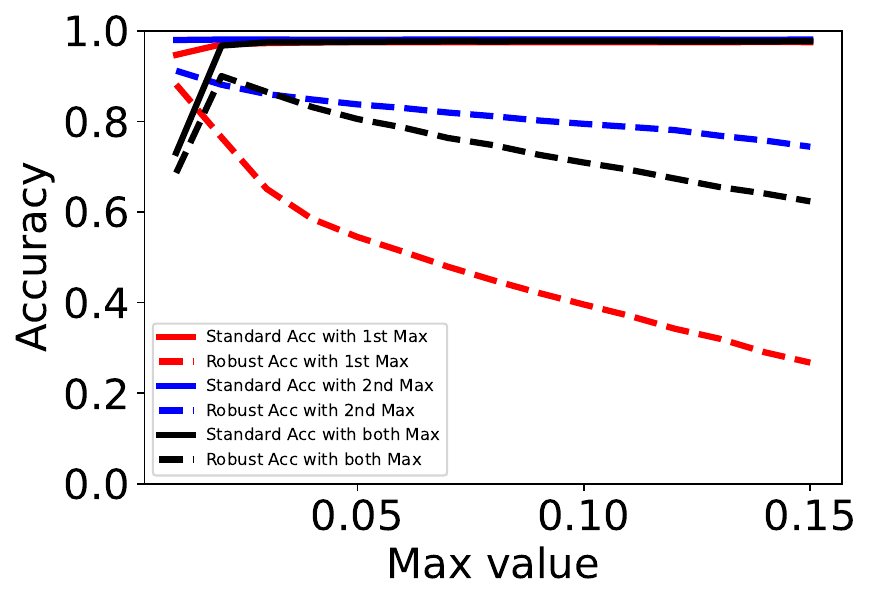}
         \label{fig:0.1_acc_two_hidden}
     \end{subfigure}
     \hfill
     \begin{subfigure}[b]{0.3\textwidth}
         \centering
         \includegraphics[width=\textwidth]{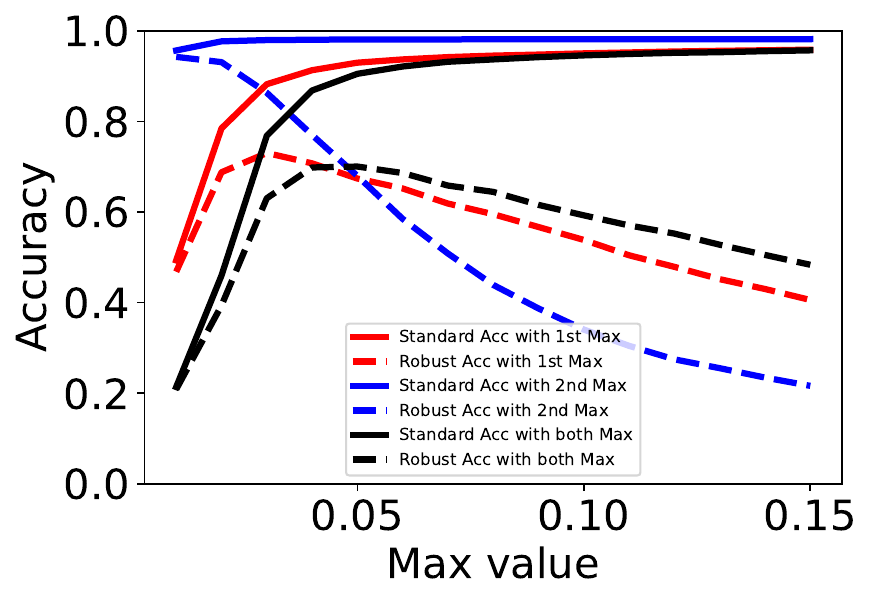}
         \label{fig:1_acc_two_hidden}
     \end{subfigure}
     \hfill     
     \begin{subfigure}[b]{0.3\textwidth}
         \centering
         \includegraphics[width=\textwidth]{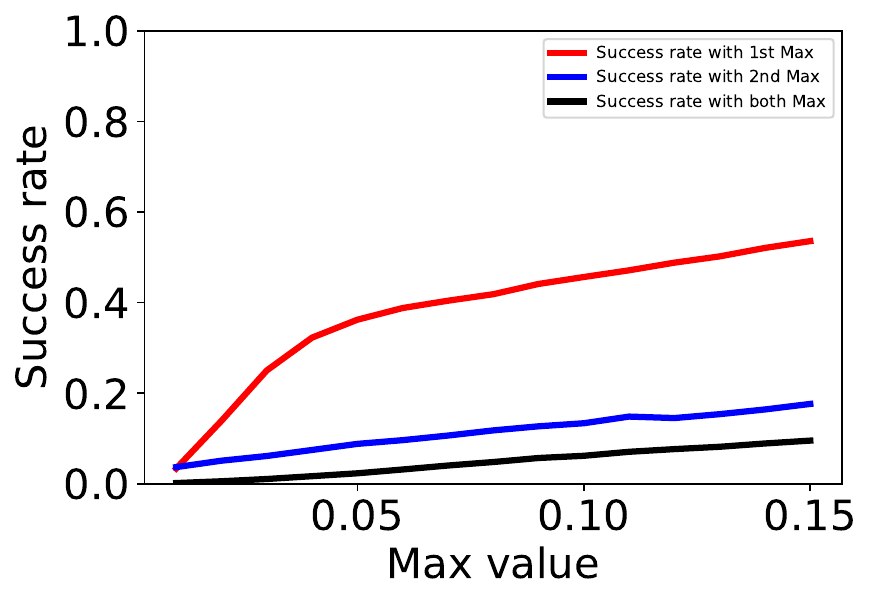}
         \caption{Initial max = $0.01$.}
         \label{fig:0.01_acc_sr_two_hidden}
     \end{subfigure}
     \hfill
     \begin{subfigure}[b]{0.3\textwidth}
         \centering
         \includegraphics[width=\textwidth]{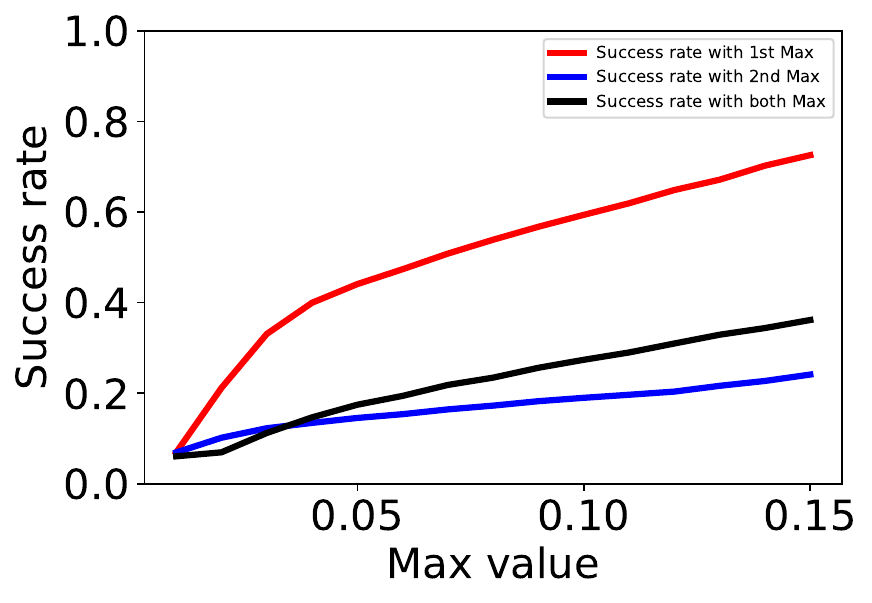}
         \caption{Initial max = $0.1$.}
         \label{fig:0.1_acc_sr_two_hidden}
     \end{subfigure}
     \hfill
     \begin{subfigure}[b]{0.3\textwidth}
         \centering
         \includegraphics[width=\textwidth]{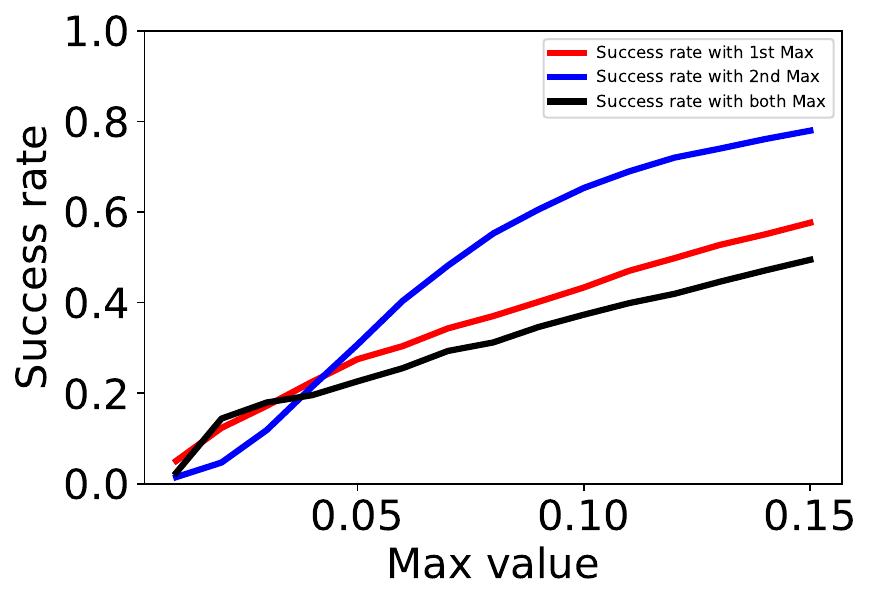}
         \caption{Initial max = $1$.}
         \label{fig:1_acc_sr_two_hidden}
     \end{subfigure}
        \caption{Standard accuracy, robust accuracy and success rate of the two-hidden-layer classifier according to PGD attack over a range of max values.}
        \label{fig:acc_sr_two_hidden}
\end{figure*}

Figure~\ref{fig:acc_sr_two_hidden} shows the result of this experiment with the general network. It demonstrates that with the initial max values of $0.01$ and $0.1$, capping the second hidden layer surprisingly outperforms the first hidden layer and capping both the hidden layers. However, with the initial max value of $1$, capping the second hidden layer underperforms the others when the max value is greater than $0.05$ in terms of robustness. Nonetheless, the robustness of the classifier being capped at the second hidden layer with a max value less than $0.05$ is higher than the others in all the max values. Therefore, capping the second hidden layer is the best solution for this general network. Because the second hidden layer has lower neurons than the first hidden layer, capping the small layer is better than capping the large layer. However, this result is derived from a specific network. Next, we experiment with the reversed network.

\begin{figure*}[h!]
     \centering
     \begin{subfigure}[b]{0.3\textwidth}
         \centering
         \includegraphics[width=\textwidth]{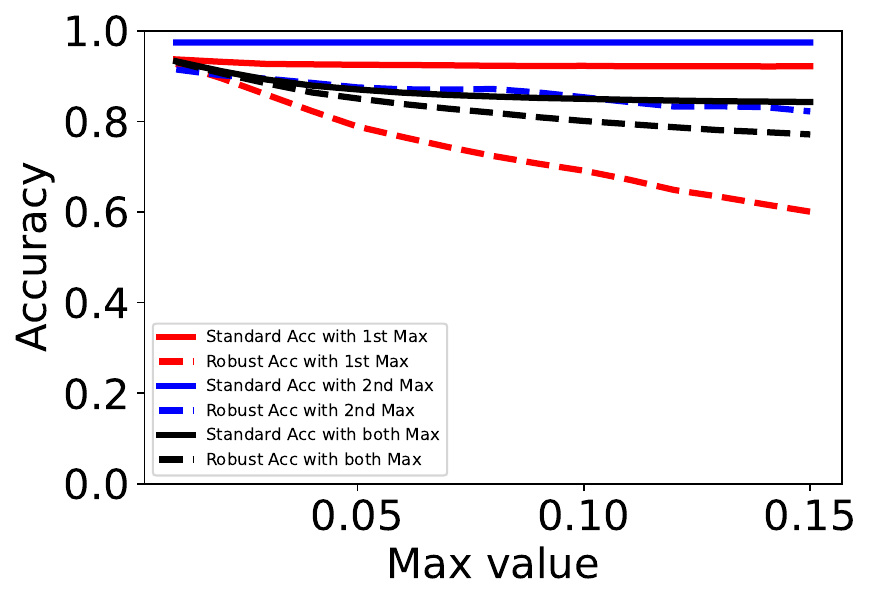}
         \label{fig:0.01_acc_two_hidden_reverse}
     \end{subfigure}
     \hfill
     \begin{subfigure}[b]{0.3\textwidth}
         \centering
         \includegraphics[width=\textwidth]{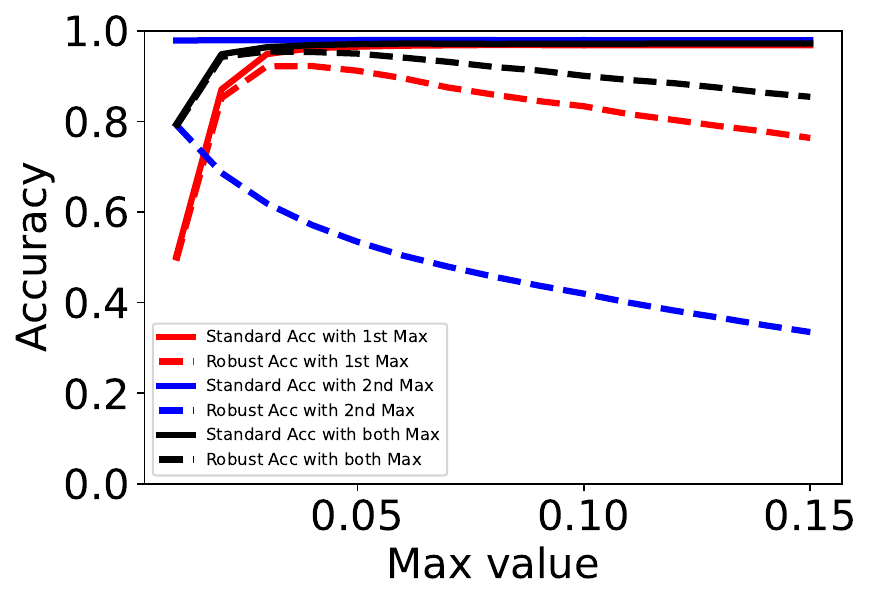}
         \label{fig:0.1_acc_two_hidden_reverse}
     \end{subfigure}
     \hfill
     \begin{subfigure}[b]{0.3\textwidth}
         \centering
         \includegraphics[width=\textwidth]{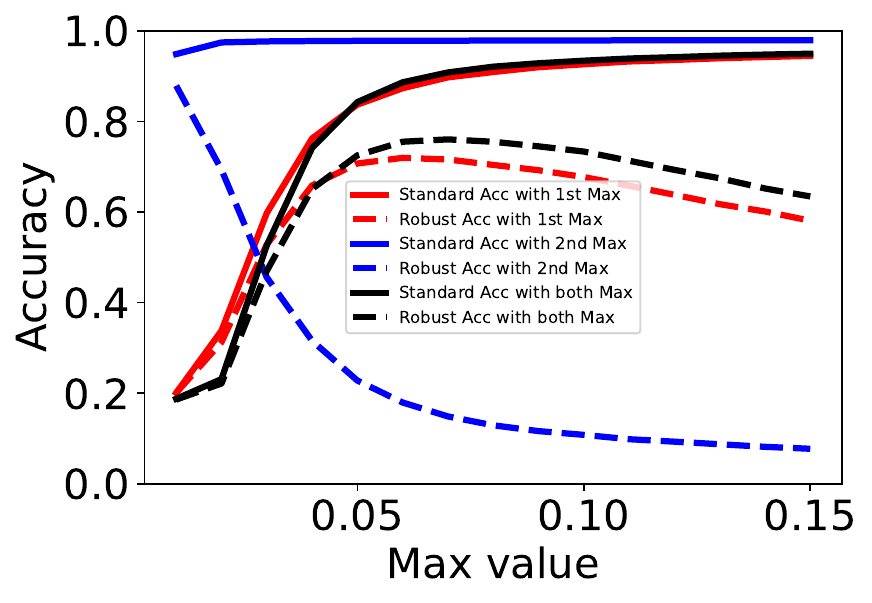}
         \label{fig:1_acc_two_hidden_reverse}
     \end{subfigure}
     \hfill     
        \begin{subfigure}[b]{0.3\textwidth}
         \centering
         \includegraphics[width=\textwidth]{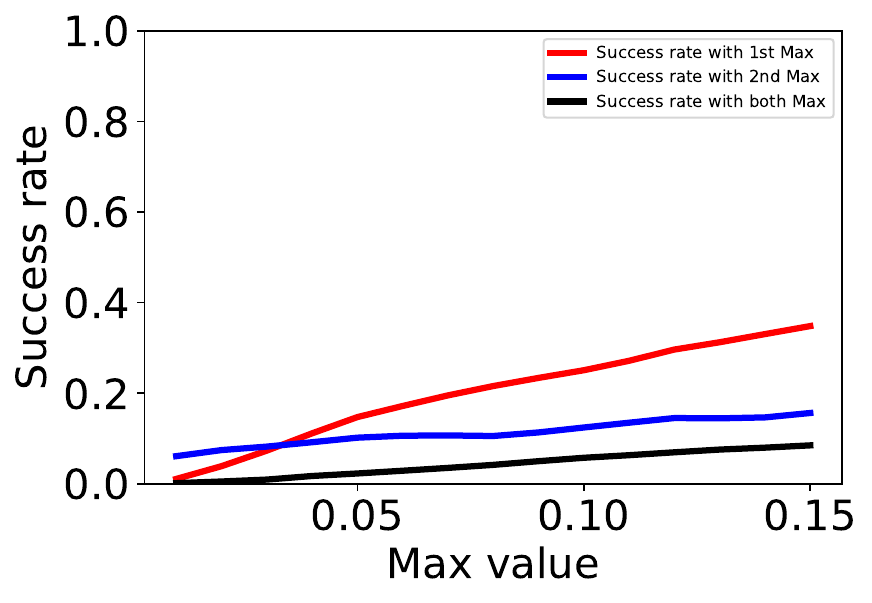}
         \caption{Initial max = $0.01$.}
         \label{fig:0.01_acc_sr_two_hidden_reverse}
     \end{subfigure}
     \hfill
     \begin{subfigure}[b]{0.3\textwidth}
         \centering
         \includegraphics[width=\textwidth]{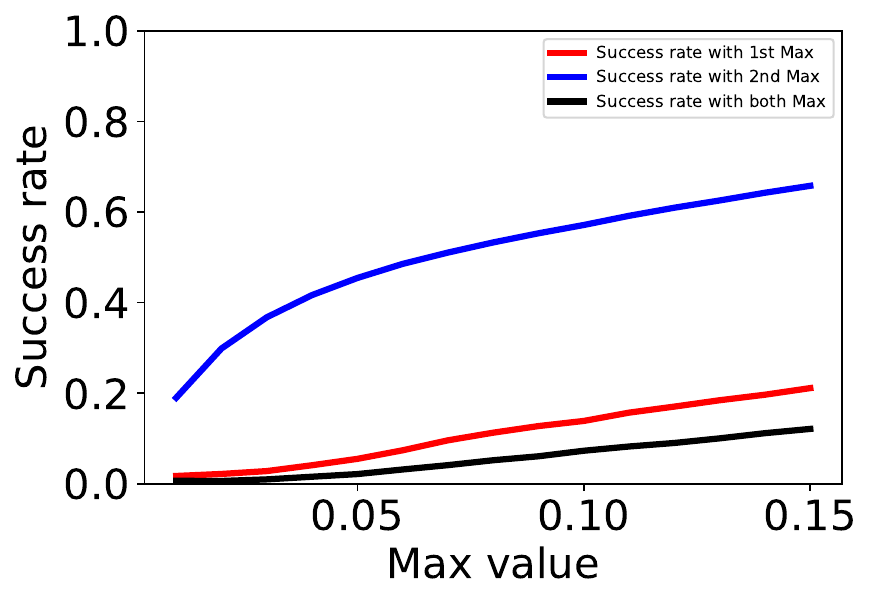}
         \caption{Initial max = $0.1$.}
         \label{fig:0.1_acc_sr_two_hidden_reverse}
     \end{subfigure}
     \hfill
     \begin{subfigure}[b]{0.3\textwidth}
         \centering
         \includegraphics[width=\textwidth]{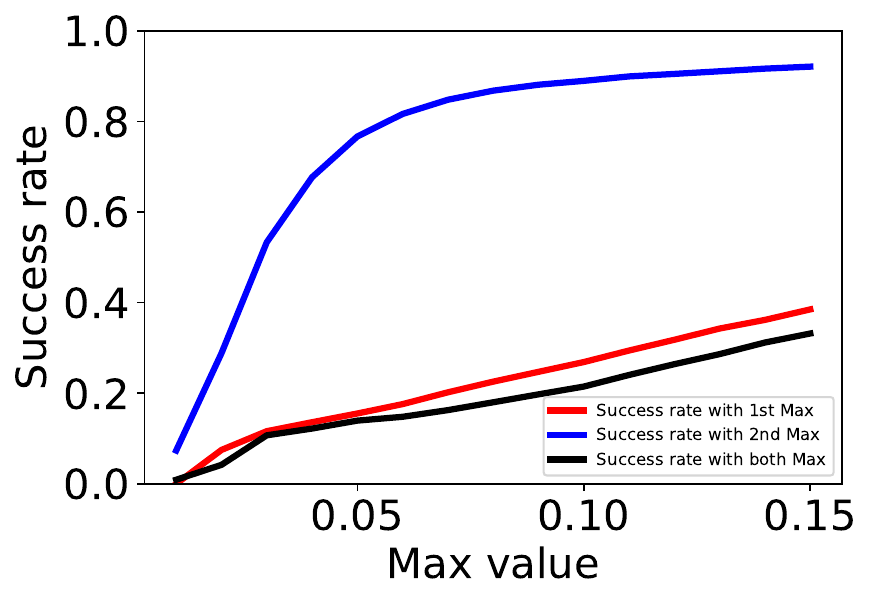}
         \caption{Initial max = $1$.}
         \label{fig:1_acc_sr_two_hidden_reverse}
     \end{subfigure}
        \caption{Standard accuracy, robust accuracy and success rate of the reversed two-hidden-layer classifier according to PGD attack over a range of max values.}
        \label{fig:acc_sr_two_hidden_reverse}
\end{figure*}

Figure~\ref{fig:acc_sr_two_hidden_reverse} shows the results of the same experiment from the reversed network. Capping both the hidden layers outperforms the others in most cases. Also, capping the first hidden layer outperforms capping the second hidden layer in most cases. Since the first hidden layer contains fewer neurons than the second hidden layer, capping a small layer is better than capping a large layer in this case. We can summarize that capping a bottleneck layer would result in the most robustness.

\section{Effect of Capped Layer's Order on Robustness \label{sec:order}}

In this section, we want to know whether capping an early or deep layer in a classifier can provide the most robustness. We conducted the same experiment as in the previous section. However, we built another classifier consisting of the input layer, two $784$-neuron hidden layers with ReLU activations, and the output layer with Softmax activation. Noticeably, the hidden layers' sizes are equal to see which layer affects the most in terms of robustness. 

\begin{figure*}[h!]
     \centering
     \begin{subfigure}[b]{0.3\textwidth}
         \centering
         \includegraphics[width=\textwidth]{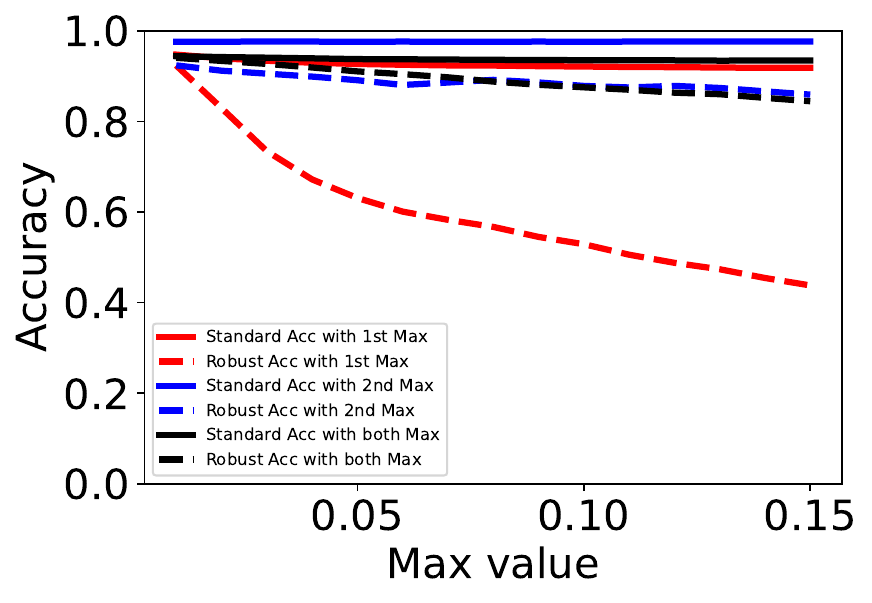}
         \label{fig:0.01_acc_two_hidden_equal}
     \end{subfigure}
     \hfill
     \begin{subfigure}[b]{0.3\textwidth}
         \centering
         \includegraphics[width=\textwidth]{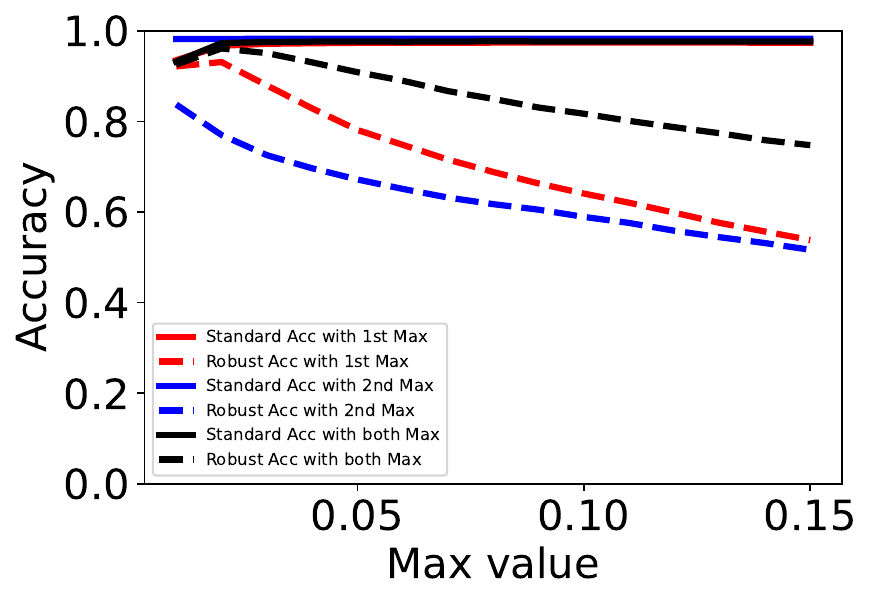}
         \label{fig:0.1_acc_two_hidden_equal}
     \end{subfigure}
     \hfill
     \begin{subfigure}[b]{0.3\textwidth}
         \centering
         \includegraphics[width=\textwidth]{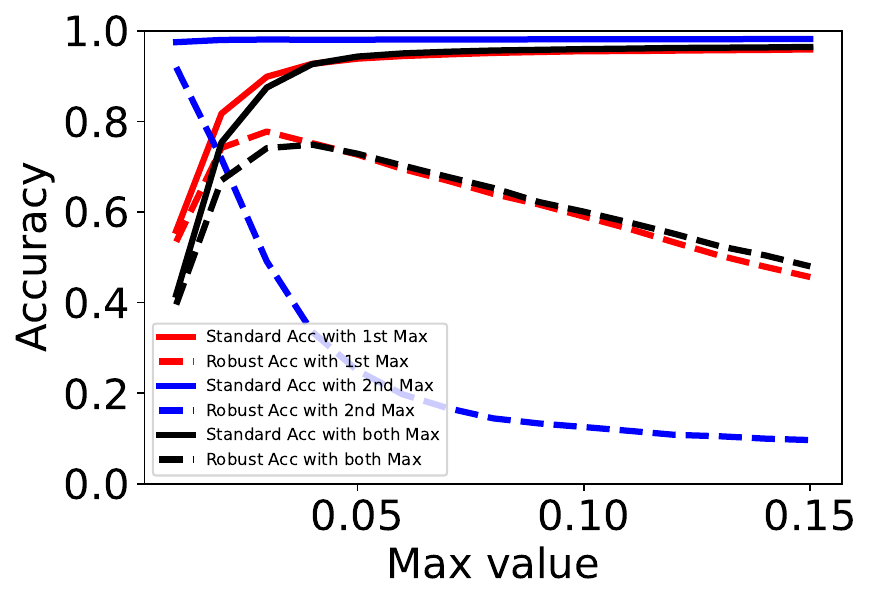}
         \label{fig:1_acc_two_hidden_equal}
     \end{subfigure}
     \hfill     
     \begin{subfigure}[b]{0.3\textwidth}
         \centering
         \includegraphics[width=\textwidth]{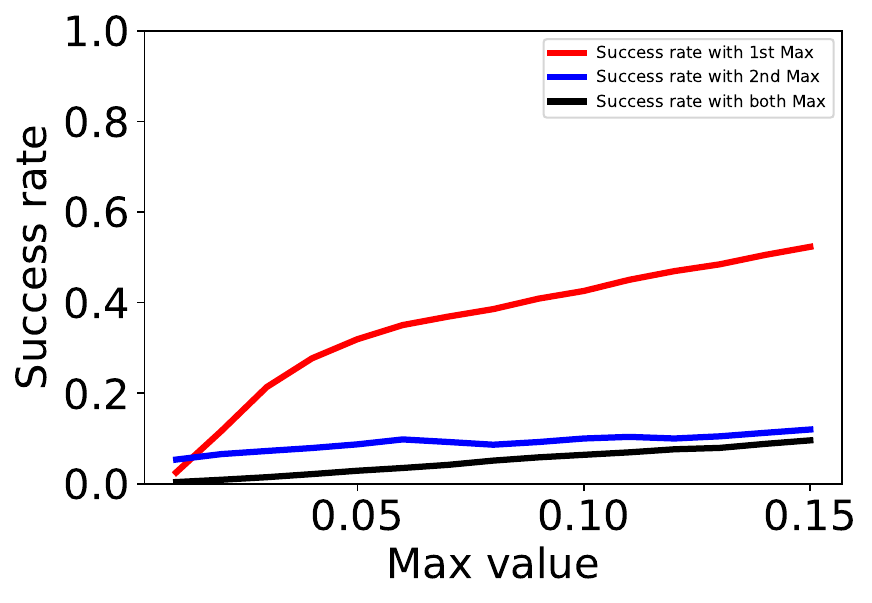}
         \caption{Initial max = $0.01$.}
         \label{fig:0.01_acc_sr_two_hidden_equal}
     \end{subfigure}
     \hfill
     \begin{subfigure}[b]{0.3\textwidth}
         \centering
         \includegraphics[width=\textwidth]{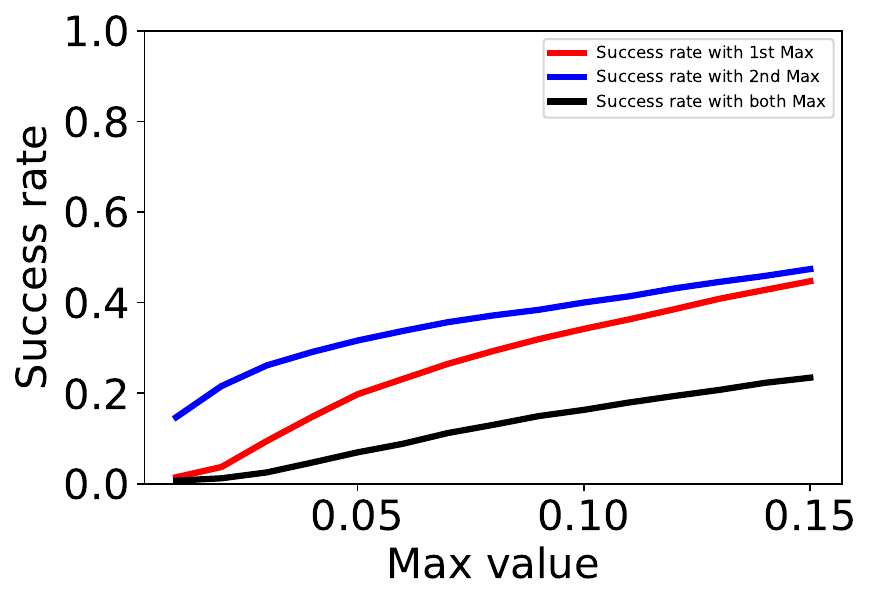}
         \caption{Initial max = $0.1$.}
         \label{fig:0.1_acc_sr_two_hidden_equal}
     \end{subfigure}
     \hfill
     \begin{subfigure}[b]{0.3\textwidth}
         \centering
         \includegraphics[width=\textwidth]{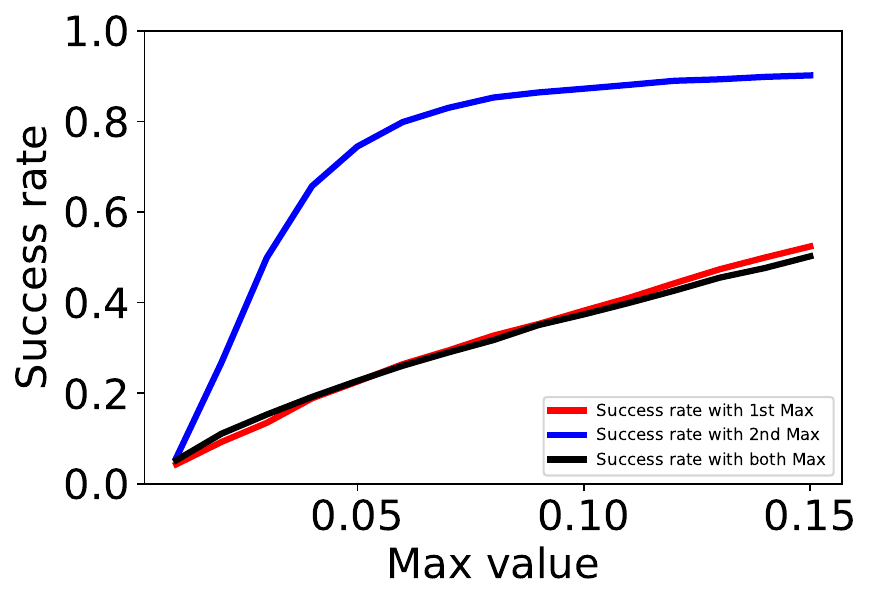}
         \caption{Initial max = $1$.}
         \label{fig:1_acc_sr_two_hidden_equal}
     \end{subfigure}
        \caption{Standard accuracy, robust accuracy and success rate of the equal two-hidden-layer classifier according to PGD attack over a range of max values.}
        \label{fig:acc_sr_two_hidden_equal}
\end{figure*}

Figure~\ref{fig:acc_sr_two_hidden_equal} shows the results of this experiment. The classifier capped at the first hidden layer performs relatively better when the initial max value grows. Evidently, at the initial max value of $0.01$, capping the second hidden layer is better than the first hidden layer. However, increasing the initial max value results in the opposite consequence. Therefore, capping the deep layer is recommended with a very low initial max value, and the early layer is preferred with a medium to high initial max value.

\section{Zero Gradient Experiment \label{sec:zero_gradient}}

\begin{figure}[h!]
    \centering
    \includegraphics[width=\columnwidth]{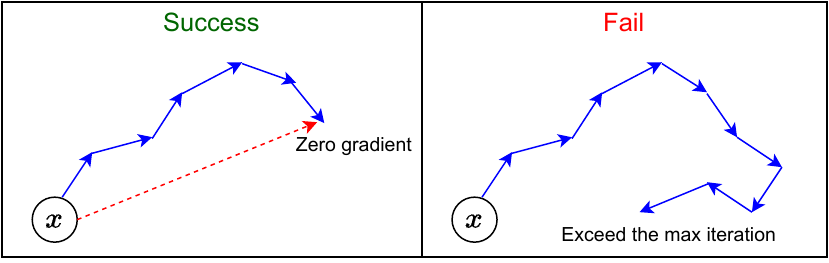}
    \caption{Examples of success and failure scenarios where a blue arrow is a gradient direction in each step of PGD attack, and the red dash arrow is the distance between sample $x$ to the zero-gradient location.}
    \label{fig:zero_gradient}
\end{figure}

In this experiment, we aim to provide further empirical evidence to support the validity of our previous results. To this end, we have modified the projected gradient descent (PGD) attack method to include a new stopping criterion, which we refer to as ``zero gradients''. Specifically, instead of terminating the attack once an adversarial example has been found, we continue the attack until the gradient of the objective function is zero. This modification allows us to assess the robustness of a targeted classifier in a more meaningful way.

We assume that a classifier is robust if the location where the zero gradients are found is close to the original, clean input sample. This is because if an attacker encounters a zero gradient, they will no longer be able to perform any gradient-based attacks, such as the fast gradient sign method (FGSM) or PGD. To measure the distance between the clean sample and the location where the zero gradients are found, we use the Euclidean distance.

However, it should be noted that this method can fail if zero gradients are not found within the maximum number of iterations. Figure~\ref{fig:zero_gradient} illustrates examples of both successful and unsuccessful scenarios. To obtain a more comprehensive understanding of the robustness of the classifiers, we compute the average distance only from the test samples where zero gradients are found. We use the MNIST dataset, along with the classifiers from Section~\ref{sec:size} and~\ref{sec:order} for this experiment.

\begin{figure*}[h!]
     \centering 
     \begin{subfigure}[b]{0.3\textwidth}
         \centering
         \includegraphics[width=\textwidth]{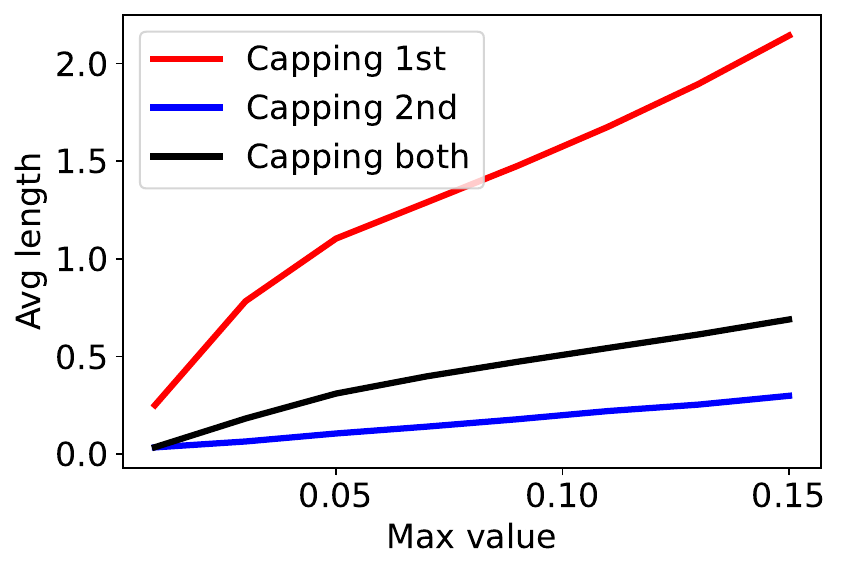}
         \caption{Initial max = $0.01$.}
         \label{fig:0.01_zero_grad_two_hidden}
     \end{subfigure}
     \hfill
     \begin{subfigure}[b]{0.3\textwidth}
         \centering
         \includegraphics[width=\textwidth]{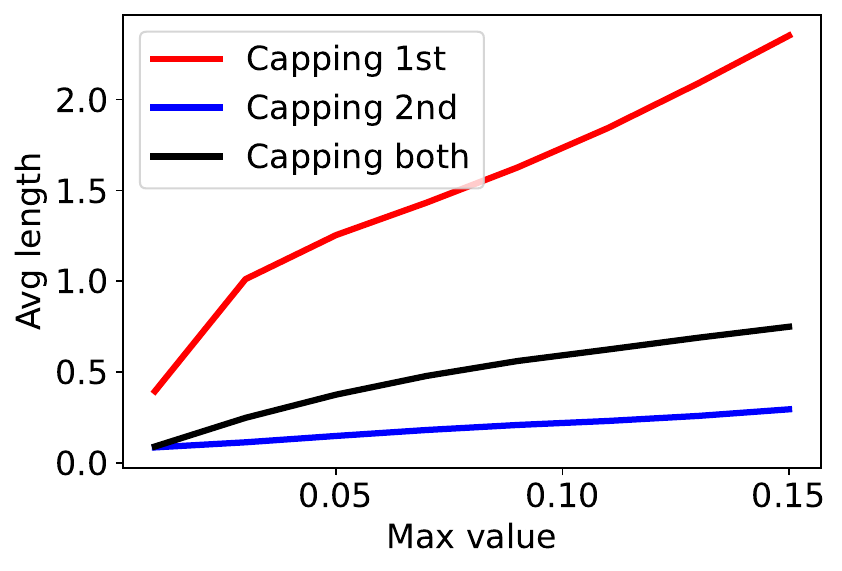}
         \caption{Initial max = $0.1$.}
         \label{fig:0.1_zero_grad_two_hidden}
     \end{subfigure}
     \hfill
     \begin{subfigure}[b]{0.3\textwidth}
         \centering
         \includegraphics[width=\textwidth]{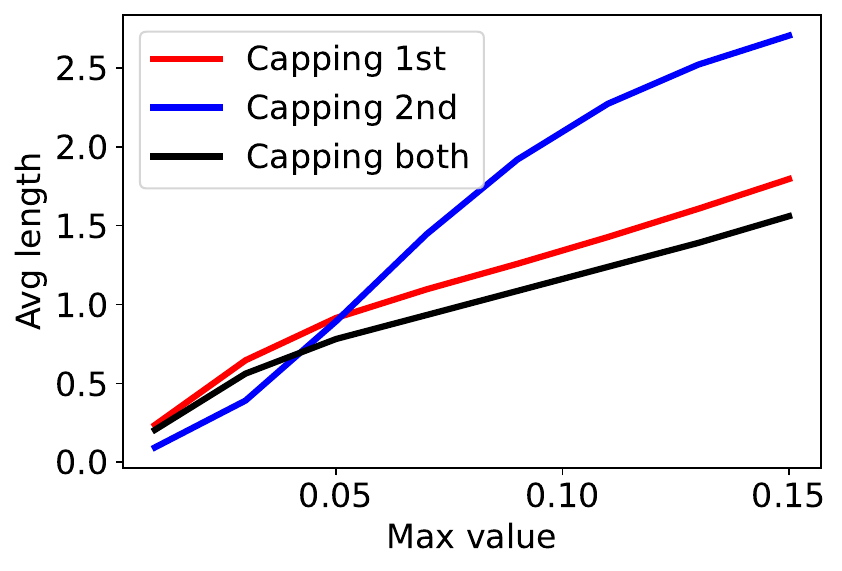}
         \caption{Initial max = $1$.}
         \label{fig:1_zero_grad_two_hidden}
     \end{subfigure}
        \caption{Average distance to zero gradients by PGD attack on a range of max values where the targets are general networks.}
        \label{fig:zero_grad_two_hidden}
\end{figure*}

\begin{figure*}[h!]
     \centering 
     \begin{subfigure}[b]{0.3\textwidth}
         \centering
         \includegraphics[width=\textwidth]{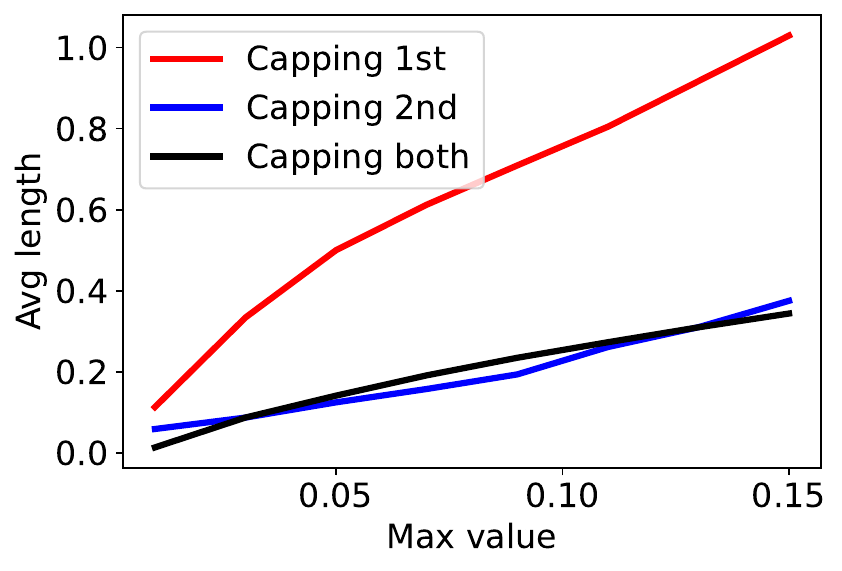}
         \caption{Initial max = $0.01$.}
         \label{fig:0.01_zero_grad_two_hidden_reverse}
     \end{subfigure}
     \hfill
     \begin{subfigure}[b]{0.3\textwidth}
         \centering
         \includegraphics[width=\textwidth]{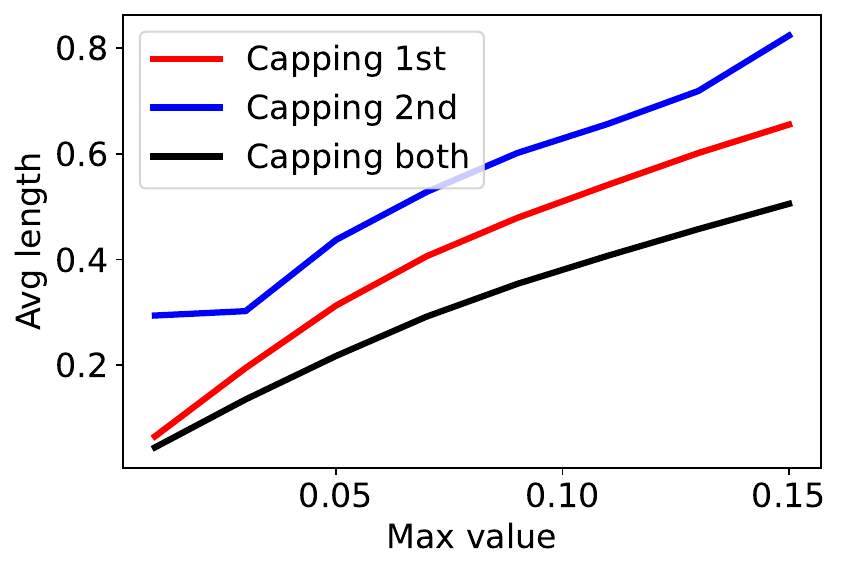}
         \caption{Initial max = $0.1$.}
         \label{fig:0.1_zero_grad_two_hidden_reverse}
     \end{subfigure}
     \hfill
     \begin{subfigure}[b]{0.3\textwidth}
         \centering
         \includegraphics[width=\textwidth]{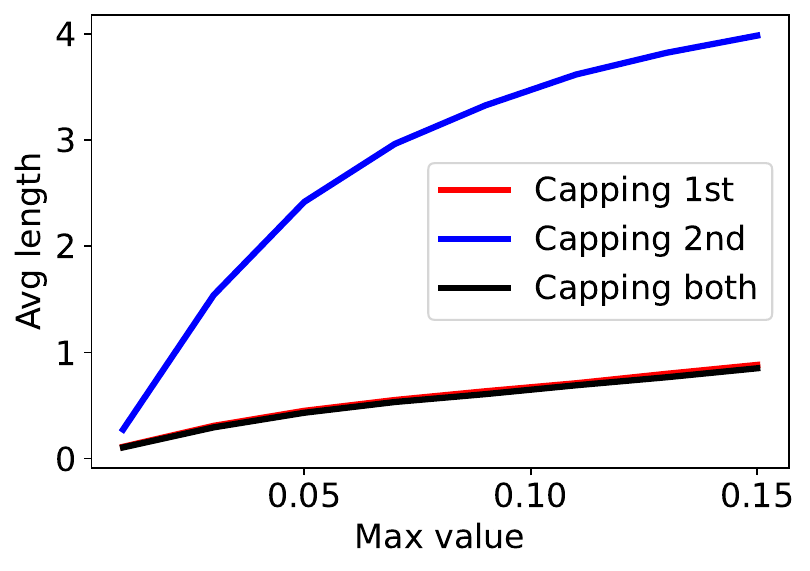}
         \caption{Initial max = $1$.}
         \label{fig:1_zero_grad_two_hidden_reverse}
     \end{subfigure}
        \caption{Average distance to zero gradients by PGD attack on a range of max values where the targets are reversed networks.}
        \label{fig:zero_grad_two_hidden_reverse}
\end{figure*}

\begin{figure*}[h!]
     \centering 
     \begin{subfigure}[b]{0.3\textwidth}
         \centering
         \includegraphics[width=\textwidth]{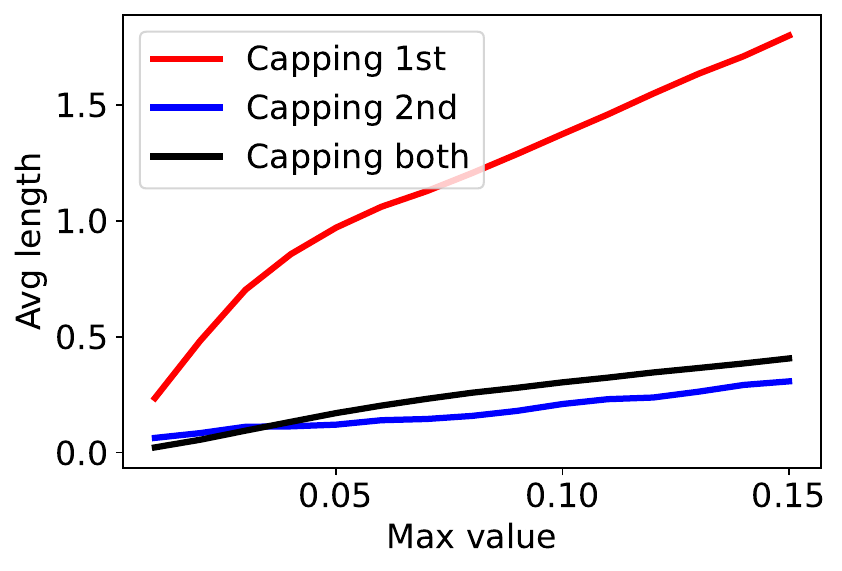}
         \caption{Initial max = $0.01$.}
         \label{fig:0.01_zero_grad_two_hidden_equal}
     \end{subfigure}
     \hfill
     \begin{subfigure}[b]{0.3\textwidth}
         \centering
         \includegraphics[width=\textwidth]{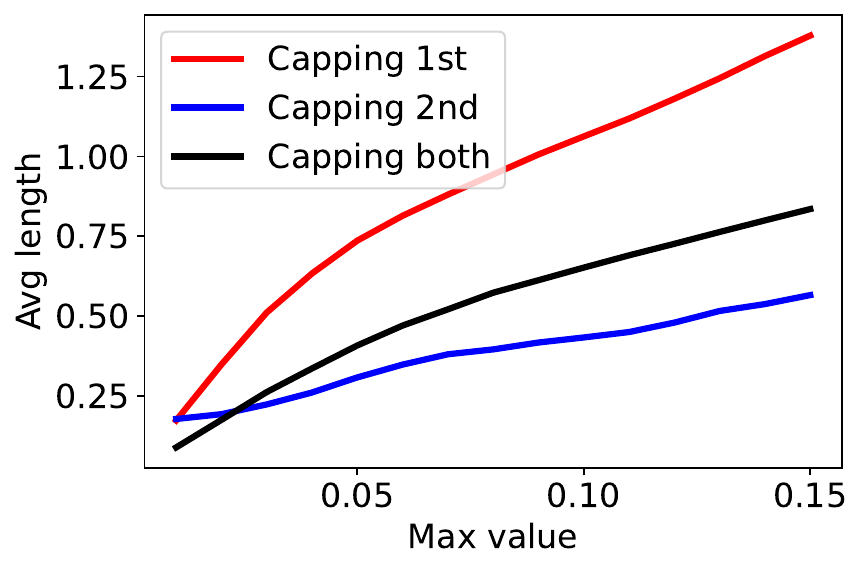}
         \caption{Initial max = $0.1$.}
         \label{fig:0.1_zero_grad_two_hidden_equal}
     \end{subfigure}
     \hfill
     \begin{subfigure}[b]{0.3\textwidth}
         \centering
         \includegraphics[width=\textwidth]{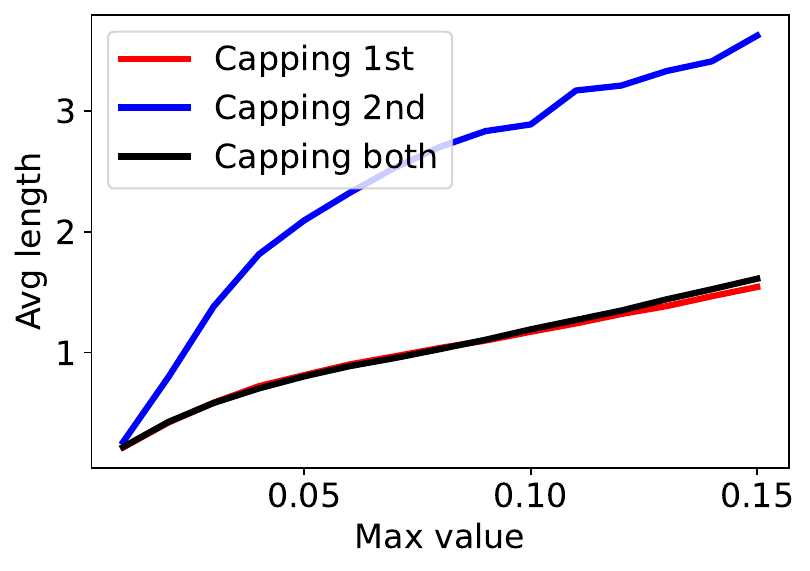}
         \caption{Initial max = $1$.}
         \label{fig:1_zero_grad_two_hidden_equal}
     \end{subfigure}
        \caption{Average distance to zero gradients by PGD attack on a range of max values where the targets are equal networks.}
        \label{fig:zero_grad_two_hidden_equal}
\end{figure*}

Figures~\ref{fig:zero_grad_two_hidden}, \ref{fig:zero_grad_two_hidden_reverse}, and~\ref{fig:zero_grad_two_hidden_equal} show the results of the zero gradient experiment with the general, reversed and equal networks, respectively. Interestingly, we found that these results aligned with the previous experiments in Section~\ref{sec:size} and~\ref{sec:order}. Therefore, we can conclude that the results in those sections are valid.

\section{Capped-ReLU Classifier's Sensitivity Map \label{sec:map}}
As discussed in~\cite{sooksatra2022evaluation}, a pixel in an image vulnerable to adversarial attacks is sensitive to a slight change. It also proposed an equation to compute a sensitivity map to determine how much each pixel is susceptible to adversarial attacks. The equation is
\begin{equation}
    \text{smap}(\bm{x}, Z) = \text{max}\left(\bm{0}, \frac{\partial Z_t}{\partial \bm{x}} \cdot \sum_{c \neq t}\frac{\partial Z_c}{\partial \bm{x}}\right),
\end{equation}
where $\bm{x}$ is an input, $Z$ is a classifier whose output is before Softmax function, $Z_i$ is the output of class $i$, $t$ is the true class of $\bm{x}$ and $\bm{0}$ is a matrix of 0 whose size is the same as $\bm{x}$. We sum the map's values across all pixels to show that capping ReLU functions improves robustness.

\begin{figure*}[h!]
     \centering 
     \begin{subfigure}[b]{0.2\textwidth}
         \centering
         \includegraphics[width=\textwidth]{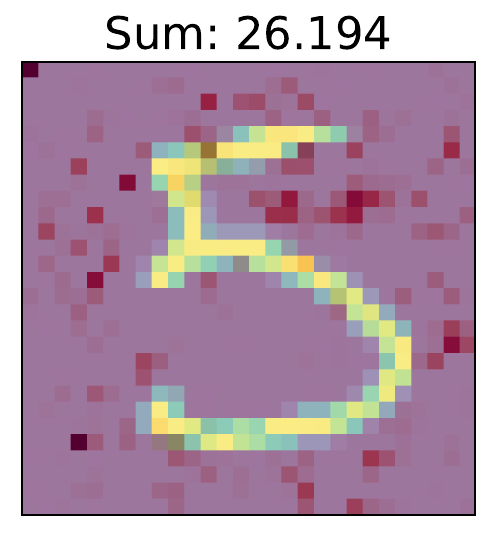}
         \caption{General ReLU.}
         \label{fig:sense_none}
     \end{subfigure}
     \hfill
     \begin{subfigure}[b]{0.2\textwidth}
         \centering
         \includegraphics[width=\textwidth]{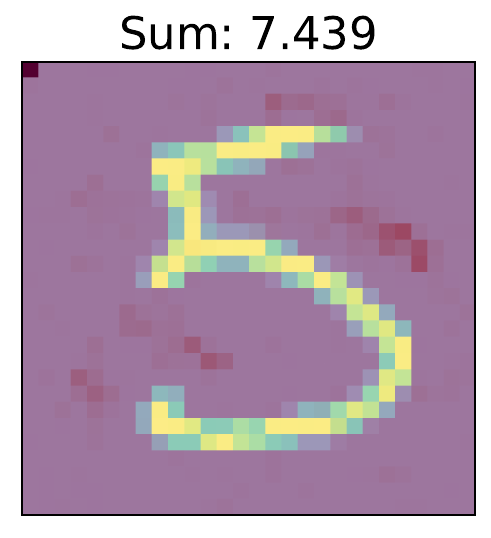}
         \caption{Max val = $1$.}
         \label{fig:sense_1}
     \end{subfigure}
     \hfill
     \begin{subfigure}[b]{0.2\textwidth}
         \centering
         \includegraphics[width=\textwidth]{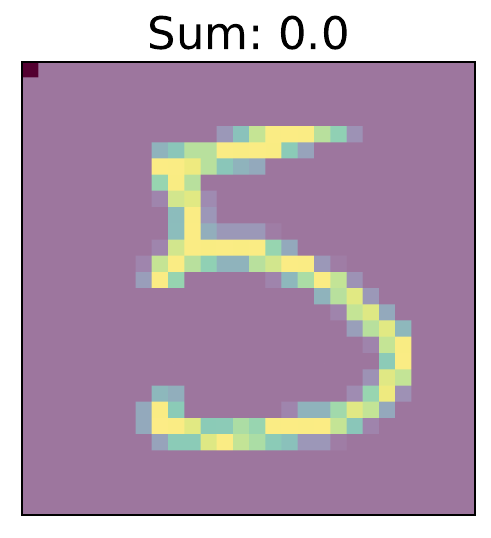}
         \caption{Max val = $0.1$.}
         \label{fig:sense_0.1}
     \end{subfigure}
     \hfill
     \begin{subfigure}[b]{0.2\textwidth}
         \centering
         \includegraphics[width=\textwidth]{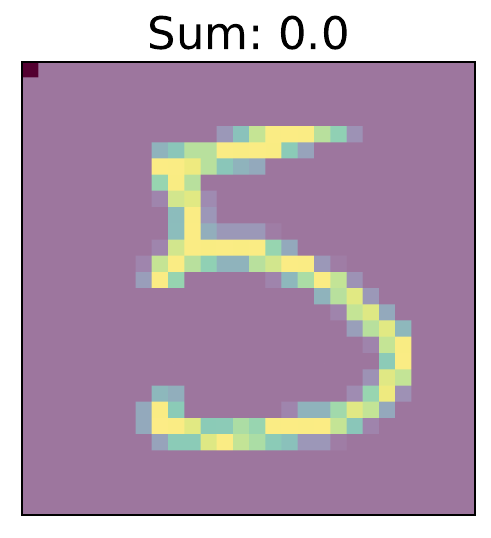}
         \caption{Max val = $0.01$.}
         \label{fig:sense_0.01}
     \end{subfigure}
        \caption{Sensitivity map of digit five and the summation of the scores on the top. Note that the more red pixel is, the more sensitive pixel becomes. Also, the black pixel in the top left of the image is not included in the map. We use it as a maximum reference value to tune the value's range across all the images.}
        \label{fig:sense}
\end{figure*}

We create a two-hidden-layer classifier and train it with several max values (i.e., $1$, $0.1$ and $0.01$). Figure~\ref{fig:sense} shows the sensitivity map of digit five with the classifier. Essentially, the number of vulnerable pixels and the summation of the map decrease when the max value is reduced. Therefore, capping ReLU functions with low max values can improve the robustness.

\section{Capped ReLU with Adversarial Training \label{sec:adversarial_training}}
In this section, we aim to investigate the efficacy of applying adversarial training techniques to capped-ReLU classifiers to enhance their robustness beyond that of general classifiers that have undergone adversarial training. To accomplish this, we utilize two-hidden-layer neural networks as the base model and train them using clean test samples for a total of twenty epochs with the Adam optimizer as described in~\cite{kingma2014adam} and a learning rate of $0.001$. We only apply the ReLU function cap at the second hidden layer, as previous sections have demonstrated this to be the most effective location for such an operation.

Following the initial training phase, we then proceed to apply adversarial training to these networks through the use of either the Fast Gradient Sign Method (FGSM) or Projected Gradient Descent (PGD) for an additional ten epochs, with a perturbation bound of $0.1$. Subsequently, we evaluate these networks' accuracy on clean test samples and samples that have been attacked using FGSM, PGD, and the Carlini and Wagner (CW) attack~\cite{carlini2017towards}. For FGSM and PGD, we employ a perturbation bound of $0.1$, a maximum iteration of $10$ and a step size of $0.01$. Additionally, in the case of the CW attack, we use a maximum iteration of $10000$, a learning rate of $0.01$, an initial balancing factor of $0.001$, and $9$ adjustments of the balancing factor.

\begin{table}[h!]
    \centering
    \caption{Accuracy of MNIST two-hidden-layer classifiers with general and capped ReLU activation functions on clean test samples and adversarial test samples generated by using FGSM, PGD and CW.}
    \begin{tabular}{rrrrrr}
    \toprule
        Max & Adv. & Clean & FGSM & PGD & CW($L_2$)\\
       Val. & Training & \%  & \%  & \%  & \% \\
        \midrule
        -      & -    & $98.49$ & $41.77$ & $9.47$  & $0.00$ \\
        $1.00$ & -    & $98.46$ & $41.24$ & $7.45$  & $0.00$ \\
        $0.10$ & -    & $98.06$ & $68.04$ & $39.79$ & $5.56$ \\
        $0.01$ & -    & $97.88$ & $92.37$ & $89.61$ & $8.07$ \\
        -      & FGSM & $98.26$ & $91.44$ & $85.12$ & $0.19$ \\
        $1.00$ & FGSM & $98.35$ & $92.46$ & $81.88$ & $0.18$ \\
        $0.10$ & FGSM & $98.18$ & $93.00$ & $90.37$ & $3.50$ \\
        $0.01$ & FGSM & $97.10$ & $94.07$ & $96.36$ & $8.21$ \\
        -      & PGD  & $98.67$ & $91.85$ & $86.74$ & $0.10$ \\
        $1.00$ & PGD  & $98.49$ & $93.32$ & $87.09$ & $0.11$ \\
        $0.10$ & PGD  & $98.09$ & $92.64$ & $92.85$ & $3.62$ \\
        $0.01$ & PGD  & $96.55$ & $89.21$ & $95.43$ & $8.00$ \\
        \bottomrule
    \end{tabular}    
    \label{tab:mnist_accuracy}
\end{table}

The configurations of the classifiers and their corresponding accuracy on both clean and adversarial test samples are presented in Table~\ref{tab:mnist_accuracy}. The results reveal that by decreasing the maximum value, the robustness of the classifiers against attacks using FGSM, PGD, and CW can be improved without sacrificing a significant portion of standard accuracy. This is particularly evident when the models are retrained using FGSM, which results in similar performance and robustness to retraining using PGD, despite the latter taking much more time, as previously discussed in~\cite{wong2020fast}. Additionally, it is worth noting that although capping the ReLU function can improve robustness, the CW attack remains particularly effective, as it is not limited by any perturbation bound. Despite the success of the CW attack, we continue to see the trend that using a lower max value yields a more robust network. Therefore, a correctly customized classifier concerning its ReLU functions would ultimately be robust against CW. In this context, it is essential to note that static capping ReLU activation functions are a starting point for enhancing adversarial robustness by customizing architecture.

\section{Conclusion \label{sec:conclusion}}
This work demonstrates that ReLU activation functions, despite their ability to increase the speed of the training process, are a significant contributor to adversarial attacks. Through experimentation, we have shown that small perturbations can rapidly grow over the layers in a classifier that utilizes general ReLU functions, ultimately leading to a change in the prediction at the output layer. However, by capping these functions with a maximum value, the growth of perturbations is controlled, improving the classifier's robustness. The zero gradient experiment and the sensitivity map support our findings. Additionally, we have established that applying adversarial training to classifiers that use capped ReLU functions can further improve robustness beyond that of classifiers that use general ReLU functions with adversarial training. These findings are of particular importance in the field of adversarial learning.

However, it is essential to note that the capping of ReLU functions also has a significant limitation. This technique creates a zero gradient area at the other end of the ReLU function, leading to a vanishing gradient problem in large architectures and negatively impacting the training process for large-scale image classifiers. This is particularly true for dense layers. On the other hand, capping ReLU functions in convolutional layers is less likely to affect the training process since it has a constant number of parameters in each filter (i.e., the filter's area times the number of input channels).

For future research, we aim to improve this approach so that classifiers can be more robust against a broader range of attacks (e.g., CW) and demonstrate scalability on large-scale classifiers. Additionally, as several large models can be imported into popular deep learning frameworks, we also aim to show that adding a layer at the beginning or end before the output layer with a customized ReLU function can control adversarial perturbations and improve robustness as an alternative to directly customizing activation functions within the models.

\bibliographystyle{icml2023}
\bibliography{references}

%%%%%%%%%%%%%%%%%%%%%%%%%%%%%%%%%%%%%%%%%%%%%%%%%%%%%%%%%%%%%%%%%%%%%%%%%%%%%%%
%%%%%%%%%%%%%%%%%%%%%%%%%%%%%%%%%%%%%%%%%%%%%%%%%%%%%%%%%%%%%%%%%%%%%%%%%%%%%%%
% APPENDIX
%%%%%%%%%%%%%%%%%%%%%%%%%%%%%%%%%%%%%%%%%%%%%%%%%%%%%%%%%%%%%%%%%%%%%%%%%%%%%%%
%%%%%%%%%%%%%%%%%%%%%%%%%%%%%%%%%%%%%%%%%%%%%%%%%%%%%%%%%%%%%%%%%%%%%%%%%%%%%%%
% \newpage
% \appendix
% \onecolumn

% You can have as much text here as you want. The main body must be at most $8$ pages long.
% For the final version, one more page can be added.
% If you want, you can use an appendix like this one, even using the one-column format.
%%%%%%%%%%%%%%%%%%%%%%%%%%%%%%%%%%%%%%%%%%%%%%%%%%%%%%%%%%%%%%%%%%%%%%%%%%%%%%%
%%%%%%%%%%%%%%%%%%%%%%%%%%%%%%%%%%%%%%%%%%%%%%%%%%%%%%%%%%%%%%%%%%%%%%%%%%%%%%%

\end{document}